\documentclass{article}[demo]

\usepackage{silence,lmodern}

\usepackage{PRIMEarxiv}

\usepackage[utf8]{inputenc} 
\usepackage[T1]{fontenc}    
\usepackage{xcolor}  
\usepackage[bookmarks=false]{hyperref}
\hypersetup{
  colorlinks=true, 
  linkcolor=red,    
  urlcolor=pink,   
  citecolor=red,    
}
\usepackage{multicol}

\usepackage{url}            
\usepackage{booktabs}       
\usepackage{amsfonts}       
\usepackage{nicefrac}       
\usepackage{lipsum}
\usepackage{fancyhdr}       
\usepackage{graphicx}       
\usepackage{multirow} 
\usepackage{amssymb}
\usepackage{pifont}
\newcommand{\xmark}{\ding{55}}%
\usepackage{boldline} 
\usepackage{tabularx}
\usepackage{setspace}
\usepackage{indentfirst}
\usepackage[font=small,labelfont=bf]{caption} 
\graphicspath{{media/}}     

\title{Smaller3d: Smaller Models for 3D Semantic Segmentation Using Minkowski Engine and Knowledge Distillation Methods
}

\author{
  Alen Adamyan \\
  \texttt{alen\_adamyan@edu.aua.am} \\
  \And
  Erik Harutyunyan\\
  erohar1311@gmail.com\\
}
\begin{document}

\maketitle

\setstretch{1.15}
\begin{abstract}
There are various optimization techniques in the realm of 3D, including point cloud-based approaches that use mesh, texture, and voxels which optimize how you store, and how do calculate in 3D. These techniques employ methods such as feed-forward networks, 3D convolutions, graph neural networks, transformers, and sparse tensors. However, the field of 3D is one of the most computationally expensive fields, and these methods have yet to achieve their full potential due to their large capacity, complexity, and computation limits. This paper proposes the application of knowledge distillation techniques, especially for sparse tensors in 3D deep learning, to reduce model sizes while maintaining performance. We analyze and purpose different loss functions, including standard methods and combinations of various losses, to simulate the performance of state-of-the-art models of different Sparse Convolutional NNs. Our experiments are done on the standard ScanNet V2 dataset, and we achieved around 2.6\% mIoU difference with a 4 times smaller model and around 8\% with a 16 times smaller model on the latest state-of-the-art spacio-temporal convents based models.

Our source code is available at:
\href{https://github.com/madanela/smaller3d}{https://github.com/madanela/smaller3d}
\end{abstract}

\setlength{\parindent}{1.5em}

\section{Introduction}

In recent years, 3D semantic segmentation has gained significant attention in computer vision due to its broad application range in various fields such as robotics, autonomous driving, medical imaging, and many more. 3D semantic segmentation aims to extract meaningful information from 3D point clouds, by label assignments of every point in the point cloud. The scene understanding of 3D with deep learning became popular by PointNet (Qi, C. R., Su, H., Mo, K., \& Guibas, L. J. (2017)) \cite{qi2017pointnet}. After that, in 2019 4d spatio-temporal convnets: MinkowskiNets(Choy, C., Gwak, J., \& Savarese, S. (2019)) \cite{choy20194d} and SparseConvNets (Graham, B., Engelcke, M., \& Van Der Maaten, L. (2018))\cite{graham20183d} were purposed which hit state-of-the-art results on release by overpassing other methods with huge margin, by suggesting to use Sparse Convolutional NNs for this task, which are working for any D dimensional tasks. This approach was quite useful because convolutional NNs are faster than other approaches like transformers, recurrent NNs, 3D convolutions and etc. These new Sparse Convolutional NNs enable to make the learning process much faster and more efficient. Studies of Sparse Neural Networks show, that in 3D sparsity of data is very vivid, and with a voxel size of 20cm around 78\% and with a voxel size of 2.5cm around 98\% of data is empty. It becomes much clear, that whenever we are using well-known 3D convolutions or any other methods that do not consider Sparsity, a lot of weights of neural networks are just learning empty space, and they discard the neural networks to learn from informative data. After that, a new technique was added on top of MinkowskiEngine, called Mix3D (Nekrasov, A., Schult, J., Litany, O., Leibe, B., \& Engelmann, F. (2021, December))\cite{nekrasov2021mix3d} which purposed using augmentation technique to mix different indoor scenes together. It proved to be much better, as this technique was working well against overfitting, and started to help models to  better understand the general context and local geometry and achieved state-of-the-art performance on the Scannet V2 dataset.

\begin{figure}
    \centering
    \includegraphics[height=60mm]{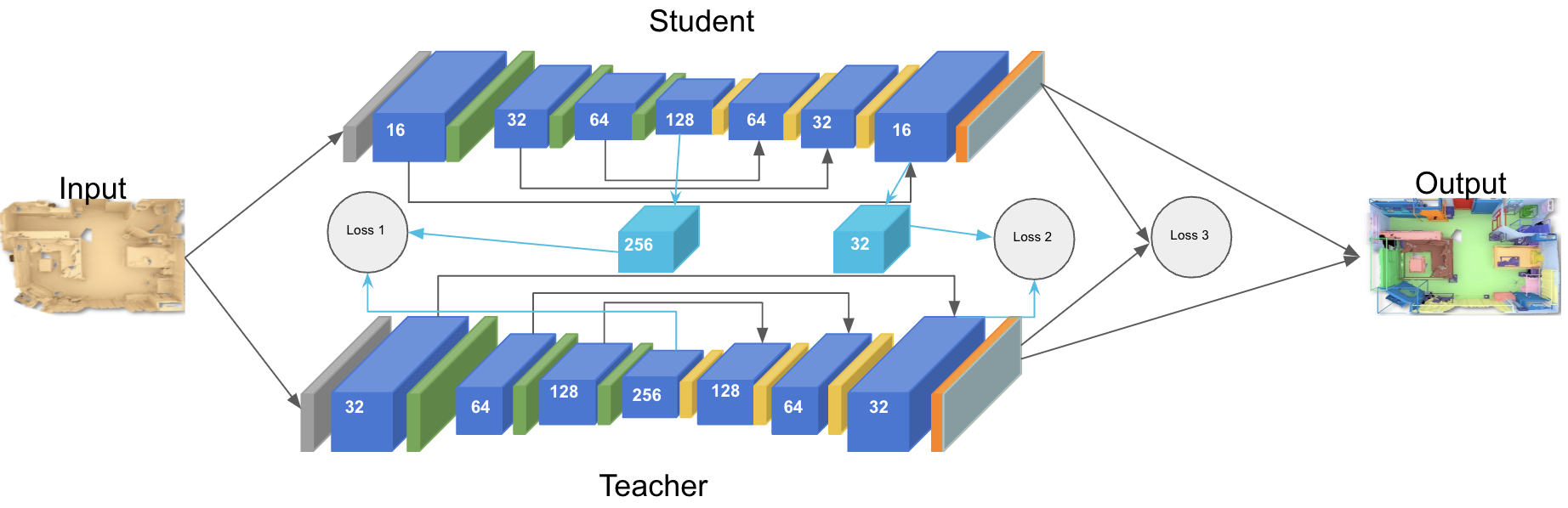}
    \captionsetup{width=.6\linewidth}
    \caption[width=4cm]{Smaller3D is a knowledge distillation method, that makes 3D Semantic Segmentation models smaller via transferring knowledge from the Teacher model to the Student model. Loss 1, Loss 2, and Loss 3 are shown as different outputs of the network, which we used to calculate the cost function. These are included in our studies to test the performance of knowledge distillation. Loss 1 is a loss between the outputs of latent spaces(Network last layer). Loss 2 is between the outputs of the U-Net similar structure Decoder feature map, and Loss 3 is between the outputs of the U-Net similar structure Encoder feature map. Note that the Convolutions shown in the visualization are 3D sparse tensors, they are cube similar for easy visualizations only.}
    \label{fig1}
\end{figure}

However, those models are still quite big and require a lot of resources (GPU, TPU) to run or test with current regular machines. In this paper, we propose the Knowledge Distillation method, which enables us to get almost the same results as state-of-the-art methods, while using much less computation and resources, and also keeping performance. We are a purposing mix of different Knowledge Distillation Loss Functions, which include using different feature map losses (See \ref{fig1}) and also Logit loss with temperature like \cite{hinton2015distilling}. Those Losses are explained in \ref{sec2}. 
In this paper, we aim to replicate the Mix3D\cite{nekrasov2021mix3d} results with a voxel size of 5cm. Then by keeping the architecture and reducing the number of neurons, use knowledge distillation to transfer teacher network info to smaller student networks. Our studies show a minimum loss in \% of mIoU(around 2\%). Through all studies, the evaluations are done on the ScanNet V2 dataset, which is concentrated on indoor 3D scans. We compare our results with different approaches, and also our own different student networks, learned by numerous loss functions. Our studies show that generalizing some tasks is done much better with a smaller model on some specific tasks, which are shown on Experiences(See \ref{exp}). We have tested our studies with 2 different networks with the structure of Res16UNet34C, by using half, or a quarter of neurons, to symmetrically reduce the sizes. We also show that this approach helps to keep the architecture and transfer original knowledge from different layers of the network, by giving us control during distillation. 
Experiments also show that using additional loss along with Knowledge Distillation standard techniques relatively speeds up learning and makes the process stable and helps student models to directly move towards teacher models and use a much higher learning rate.

\textbf{Our contributions} are summarized as:
\begin{itemize}
   \item  Apply different Knowledge Distillation methods to 3D Semantic Segmentation tasks. As a result, we achieved making relative mean intersection over union (mIoU) score, compared with the teacher network(by minimizing the network size 4 and 16 times). With this, we show that Knowledge Distillation makes training and inference complexity lower and faster while maintaining performance.
   \item Purpose a new architecture for Knowledge Distillation models for student-teacher distillation training. As a result, the student Network is able to learn from the teacher network, not only by the last layer but also through feature maps of intermediate layers.
   \item Purpose a new loss function, that combines Encoder, Decoder, and classic Knowledge Distillation losses. It gives much control over the training process, and hence with this new loss function, we achieve training process stability.
\end{itemize}

We have included all in our code and models, which were developed on top of Mix3D augmentation techniques, at \href{https://github.com/madanela/smaller3d}{https://github.com/madanela/smaller3d}

\section{Background}
\label{sec2}

The background of methods in Knowledge Distillation is very diverse, depending on the modality and scalability of models that are processed. To understand the logic of Knowledge Distillation in the 3D Semantic Segmentation task, we started by using multiple classic methods of Knowledge Distillation with a mix of different losses.

The trivial idea that can be used to train a smaller model for the same task as the teacher model, is just training it with the same pipeline, that has been done with the teacher model. So in this case, the student model does "not see", the output of the teacher model. With this approach, the labels are static, while the output of the teacher model is dynamic(See \cite{hinton2015distilling}). Firstly, tests have been done with classic Knowledge Distillation methods, which first were purposed by Bucilua, C., Caruana, R., \& Niculescu-Mizil, A. (2006, August)\cite{bucilua2006model} and then developed by Hinton, G., Vinyals, O., \& Dean, J. (2015)\cite{hinton2015distilling}. Hinton, G., Vinyals, O., \& Dean, J. (2015), suggest, that for Knowledge distillation, it is enough to use penalty only on the last layer output. It was also suggested to use temperature for "soft labels", which helps the learning process to be faster. Here on the bottom, it is represented their suggested loss, which works quite well, to imitate the teacher model performance. The gradient of the distilled model(student) is $z_i$, the temperature is T, and $v_i$ is teacher model output \cite{hinton2015distilling}.
\label{hintonformula}

$$
\frac{\partial C}{\partial z_i} =  \frac{1}{T} \left(\frac{e^{z_i/T}}{ \sum_{j} e^{z_j/T}} - \frac{e^{v_i/T}}{\sum_{j} e^{v_j/T}} \right)
$$

In that same paper, it was also suggested to use the $\alpha$ parameter between 2 different losses (student-ground truth and student-teacher) which balances loss, and by different problems, can be changed for different problems. Let in our further studies, denote this loss as $Loss_1$ as $\mathcal{L}_{BAN}$ in \ref{hintongeneralformula}.
\label{hintongeneralformula}
$$
\mathcal{L}_{BAN} = \mathcal{L}_{CE}(y,\sigma(s)) + \mathcal{L}_{KL}(\sigma(t),\sigma(s))
$$
Furthermore, we decided to add additional losses, to boost the performance of the model. In KENet (Liu, H., Zhang, C., Xu, C., \& Li, G. (2020))\cite{liu2020kenet}, they suggest to use of a basic mapping function which maps from a three-dimensional attention activation layer to 2d spatial activation map. It helps to transform both student and teacher network layerwise transformation to the same space, and then use $L_2$ loss(MSE) between pairwise subtraction of same layer matrixes.

In our case, we decided to use another approach to match different space matrixes. To do so, we suggest using Upsampling Spatio Temporal convolutional layer on top of our selected feature map, to bring it to the same space as the teacher feature map. The formula mentioned \ref{Upsample}, it is shown the $F_{UpSample}$ that is used to transform $s_{out}$(feature map output of student network) to the same space as $t_{out}$(feature map output of teacher model).

\label{Upsample}
$$
\mathcal{L}_{FM} = \mathcal{L}_{MSE}(t_{out},F_{UpSample}(\sigma(s_{out})))
$$

This Upsampling layer over the feature map is shown in Figure \ref{fig1} with Loss 2 and Loss 3 which are done over feature maps of the Decoder and Encoder last layers respectively. This method has a function of Dimensionality reduction, as with student-teacher respective layers, we force smaller dimensional feature maps of the student layer, to represent the same information as the teacher's one.

\label{sec:headings}

\section{Related Work}

In this Section, we will first introduce KD methods, or approaches, to teach smaller student models from teacher cumbersome models. Then, the related work will continue with current approaches of 3D Semantic Segmentation (describing which models exist, and which ones we tested to make smaller) and also the ones, that have been tested Knowledge Distillation in 3D. See \ref{sec:headings}.
Note, that Knowledge Distillation in 3D semantic Segmentation is not developed yet, so we decided to analyze 2 different sphere information to further mix different methods from Knowledge Distillation and Semantic Segmentation in 3D.
There are also some works done on Knowledge Distillation in 3D, but those researchers were not done on Semantic Segmentation tasks.

\subsection{Knowledge Distillation}
Knowledge Distillation(KD) is a model compression technique that started its main journey by Hinton, Vinyals, and Dean (2015)\cite{hinton2015distilling} that involves transferring knowledge from a larger, cumbersome model (teacher) to a smaller, faster model (student). The main idea is to use the teacher's output probabilities (softmax) as "soft labels", which makes the process of class learning much more efficient, by using Temperature in loss calculations. It is shown in formula \ref{hintonformula}.  The same paper also suggested using hyperparameter $\alpha$ to balance the loss between student-teacher and student-ground truth losses, to find a better loss for distillation. See formula on \ref{hintongeneralformula}.

Later Furlanello, T., Lipton, Z., Tschannen, M., Itti, L. \&amp; Anandkumar, A.. (2018) suggested Born Again Neural Networks\cite{pmlr-v80-furlanello18a} which shows, that without softening labels, it is possible to achieve same results with similar architecture. They use multiple student models, which we will discuss further, that are trying to iteratively teach student models. With each step, the student model becomes a new teacher for further distillation.

There are also early approaches, that suggested mimicking the teacher's output distribution with regularization, which is adding hint and guide layers on teacher and student models that help against overfitting. (Romero, A., Ballas, N., Kahou, S. E., Chassang, A., Gatta, C., \& Bengio, Y. 
(2014))\cite{romero2014fitnets}. It suggests classic regression layers, as additional layers, and also custom MSE loss between those layers to help mimic. Also, there are approaches, that use an idea of an additional layer adding(similar to our approach), but on a 2D Semantic Segmentation task. For example, He, T., Shen, C., Tian, Z., Gong, D., Sun, C., \& Yan, Y. (2019)\cite{he2019knowledge} suggests, using the bilinear function for upsampling, to make teacher and student networks on the same space. By our estimations, it is not relevant to our task, as they learn affinity module, which is not clear to implement with sparse tensors. Another point is, that in their tested models, the architecture is not similar to ours, and it is not relevant to the Knowledge Distillation on 3D Semantic Segmentation task.  Some other examples are shown by (Shu, C., Liu, Y., Gao, J., Xu, L., \& Shen, C. (2020).)\cite{shu2020channel}, (Wang, Y., Zhou, W., Jiang, T., Bai, X., \& Xu, Y. (2020))\cite{wang2020intra}.

Some new methods are trying to train a smaller model by using multiple teacher models. It helps inter-data-exchange over the training process between small student models, and also generalization by combining multiple methods. Also, there are approaches when multiple teacher models train smaller student models, and student model loss is decided by cross-function with weights to distill knowledge from multiple teacher models, and approximate information in students (Wang, L., \& Yoon, K. J. (2021).)\cite{wang2021knowledge}. A similar approach is suggested by different papers, which in general is designed to find an optimal number of teacher networks, and what kind of loss is needed to use, in order to make the student model less biased, and better in general. Such examples are (You, S., Xu, C., Xu, C., \& Tao, D. (2017, August))\cite{you2017learning}, (Tarvainen, A., \& Valpola, H. (2017))\cite{tarvainen2017mean}, (Liu, I. J., Peng, J., \& Schwing, A. G. (2019))\cite{liu2019knowledge}, (Zhang, Y., Xiang, T., Hospedales, T. M., \& Lu, H. (2018))\cite{zhang2018deep}, (Mirzadeh, S. I., Farajtabar, M., Li, A., Levine, N., Matsukawa, A., \& Ghasemzadeh, H. (2020))\cite{mirzadeh2020improved}, (Chen, D., Mei, J. P., Wang, C., Feng, Y., \& Chen, C. (2020, April))\cite{chen2020online}, (Park, S., \& Kwak, N. (2019))\cite{park2019feed}. In KENet (Liu, H., Zhang, C., Xu, C., \& Li, G. (2020))\cite{liu2020kenet} there is a similar idea, that we used, which is loss over feature map. However, their loss function is based on attention map distances between students and teachers in some feature map layers. However, as their Degree of Freedom is different, the distance is rather an approximation, but regular $L_1$ or $L_2$ losses.

These are also special frameworks, designed for Knowledge Distillation for neural network architectures, which helps to speed up the process of architectural design, data loading, and parallel training of student teachers (maybe teachers), and also to make the process faster for research. It is mainly using config files, to call build-in functions(mainly yaml config files). It enables to easy use of state-of-the-art Knowledge Distillation methods, for analyzing different models.
(Matsubara, Y. (2021, May))\cite{matsubara2021torchdistill}. In our case, this is quite difficult, because we are using Minkowski Engine (Choy, C., Gwak, J., \& Savarese, S. (2019).)\cite{choy20194d}, which itself is a built-in C++ library, and is not easy to integrate to any torch standard frameworks.

\subsection{Knowledge Distillation in 3D}

Knowledge Distillation in 3D is not developed as a holistic approach, as there are different evaluation benchmarks, such as Scannet V2 (Dai, A., Chang, A. X., Savva, M., Halber, M., Funkhouser, T., \& Nießner, M. (2017))\cite{dai2017scannet}, S3DIS(Armeni, I., Sener, O., Zamir, A. R., Jiang, H., Brilakis, I., Fischer, M., \& Savarese, S. (2016))\cite{armeni20163d}, SemanticKITTI(Behley, J., Garbade, M., Milioto, A., Quenzel, J., Behnke, S., Stachniss, C., \& Gall, J. (2019))\cite{behley2019semantickitti}, nuScenes(Caesar, H., Bankiti, V., Lang, A. H., Vora, S., Liong, V. E., Xu, Q., ... \& Beijbom, O. (2020))\cite{caesar2020nuscenes}. One of the main features of current Knowledge Distillation methods in 3D is that they focus mainly on either object detection or lidar Semantic Segmentation. Those methods mainly focus on outside datasets for self-driving applications. On the other hand, the Scannet V2 dataset focuses on indoor scans. Those differences are crucial, as they make a difference when using some specific methods (e.g. Augmentation techniques like Mix3D\cite{nekrasov2021mix3d}).

Lidar Semantic Segmentation is an area that has seen significant advancement in 3D knowledge distillation. One of the examples is Liu, Y., Chen, K., Liu, C., Qin, Z., Luo, Z., \& Wang, J. (2019)\cite{liu2019structured}, that suggest simultaneously distilling knowledge at three different levels: pairwise similarity, pixel-level, and holistic. Moreover, it is well introduced in the "Point-to-voxel knowledge distillation for lidar semantic segmentation" paper by
Hou, Y., Zhu, X., Ma, Y., Loy, C. C., \& Li, Y. (2022) \cite{hou2022point}, which suggests using point-to-voxel knowledge distillation approach for lidar Semantic Segmentation. Also, there is an approach which is targeting object detection tasks in 3D but is also able to transfer its knowledge to 3D Semantic Segmentation. This is a paper suggested by Yang, J., Shi, S., Ding, R., Wang, Z., \& Qi, X. (2022)\cite{yang2022towards}, which suggests its 3D KD pipeline, but their suggested pipeline performance is not good on the Semantic Segmentation tasks.

There is a wide range of Attention-based Knowledge Distillation algorithms, that are working on Semantic Segmentation tasks too. However, those architectural solutions are not relevant to sparse matrixes. Such examples of methods are (Jiang, F., Gao, H., Qiu, S., Zhang, H., Wan, R., \& Pu, J. (2023))\cite{jiang2023knowledge},(ao, Y., Zhang, Y., Yin, Z., Luo, J., Ouyang, W., \& Huang, X. (2022))\cite{yao20223d}. Their evaluation tasks are done mainly by comparing with their relative Transformers or very earlier methods. Thus, those methods are irrelevant to Convolutional networks and are not promising in this case.

One of the recent studies was 3D Point Cloud Pre-training with Knowledge Distillation from 2D Images done by ao, Y., Zhang, Y., Yin, Z., Luo, J., Ouyang, W., \& Huang, X. (2022)\cite{yao20223d}. They try to use 2D image knowledge and distill information to the student model, which has a different architecture and is designed for 3D Point cloud reconstruction. They evaluate the performance of the model with mainly object detection on Scannet V2 and also Semantic Segmentation in 3D as well. They compare their results with similar task models, and their results are around ~46\% mIoU which is 20\% off from Mix3D\cite{nekrasov2021mix3d} or Minkowski engine\cite{choy20194d} methods. 

\section{Method}
We present the Knowledge Distillation technique for 3D deep learning, aiming to make current state-of-the-art methods smaller while maintaining their performance. With this approach, we make current models computationally less expensive, and faster during inference.

In Smaller3d, we use numerous classic, and a mix of classic methods, to analyze and get a better version of Knowledge Distillation for the 3D Semantic Segmentation task. We have reduced the number of neurons on every layer of some top-ranking model architectures like (Minkowski Engine\cite{choy20194d} and Mix3D\cite{nekrasov2021mix3d}) and approximate results while making them much smaller. By testing with versions Half and Quarter, we reduce the number of neurons from $N$ to $N/4$ and $N/16$ respectively, which makes a huge difference in required resources and faster inference.
\subsection{Model}
\label{model}

For the Knowledge Distillation task in 3D Semantic Segmentation, we took Minkowski\cite{choy20194d} networks architecture as a base model. In particular, we selected standard Res16UNet34C suggested in MinkowskiNets(Choy, C., Gwak, J., \& Savarese, S. (2019)) \cite{choy20194d}, which then was used by Mix3D\cite{nekrasov2021mix3d} to run augmentation during the pipeline, that achieved state-of-the-art result on Scannet V2 benchmark with test mIoU(78.1\%) and voxel size 2cm. This architecture has U-NET similar structure and is implemented with sparse tensors while keeping the logic of U-NET Encoder-Decoder architecture. One example of those architectures is shown in Figure bottom(see \ref{figU-net}), which we used for our experiment. The one shown in the figure is {$Res16UNet34C_{Half}$, which we are talking about later in this section.

\begin{figure}[hbt!]
    \centering
    \includegraphics[height=50mm]{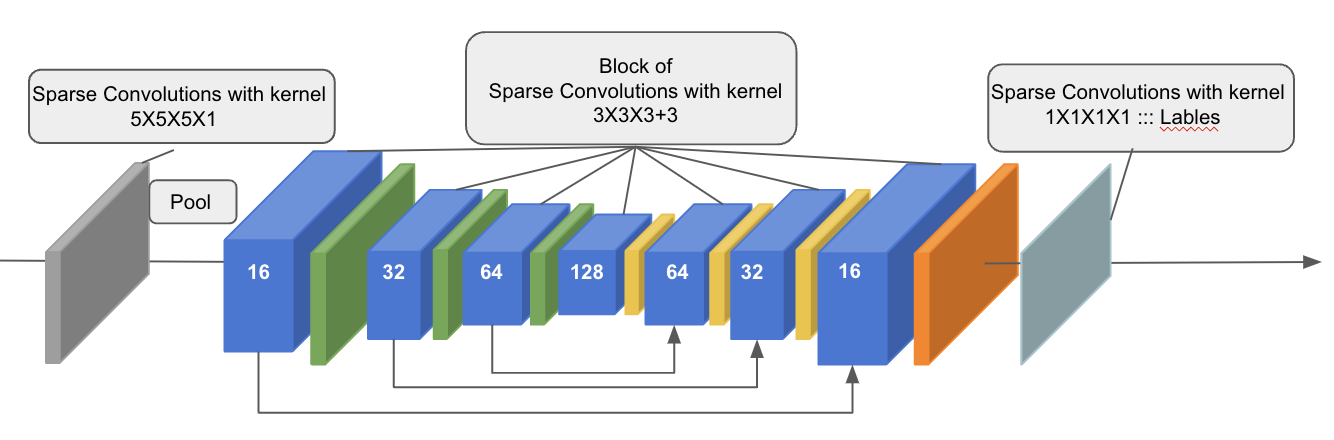}
    \captionsetup{width=.6\linewidth}
    \caption[width=4cm]{$Res16UNet34C_{Half}$, which is similar architecture, like Res16UNet34C but with half neurons on every layer. The reason we did not use the Res8UNet34C name, is to make the naming much clear for the Knowledge Distillation task. However, this name can be used for further analysis.} 
    \label{figU-net}
\end{figure}

During this step, we have 2 new models, named $Res16Net34C_{Half}$ and $Res16Net34C_{Quarter}$. In the original architecture of Res16Net34C\cite{choy20194d}, it is using (32, 64, 128, 256, 256, 128, 96, 96) number of kernels in each block of Sparse Convolutional layers. In our experiments, we have used this same architecture based on implementations of MinkowskiNets\cite{choy20194d} and applied standard Res16Net34C to replicate the results of Mix3D\cite{nekrasov2021mix3d} with their augmentations. Then, for student networks, we purposed $Res16Net34C_{Half}$ with (16, 32, 64, 128, 128, 64, 48, 48) and $Res16Net34C_{Quarter}$ with (8, 16, 32, 64, 64, 32, 24, 24)
number of kernels for each block of Sparse Convolutions. During our analysis, we mainly focused on $Res16Net34C_{Half}$. We show during our analysis, that this structure is capable of similar results, as the teacher $Res16Net34C$ network. By reducing the number of neurons on each block, it is possible to approximate the teacher model with the student model by 4 times reduction on the network number of parameters with $Res16Net34C_{Half}$ and relatively low approximation with 16 times smaller $Res16Net34C_{Quarter}$ network.
\subsection{Loss}
\label{loss}
In this subsection, we are going to describe different loss functions we used, to test $Res16Net34C_{Half}$ and $Res16Net34C_{Quarter}$ models to distill knowledge from $Res16Net34C$ model learned with Mix3D\cite{nekrasov2021mix3d} augmentation technique. For the background of our losses, we See \ref{sec2}.

For Loss analysis, we are using initially classic loss functions. Firstly, we used full loss with temperature on logits of student-teacher networks (See \ref{hintongeneralformula}) as Knowledge Distillation initial method. By using a comparison of the most trivial methods, to complicated ones. To do so, we first took $\alpha = 1$, as it makes the Knowledge Divergence Loss, suggested by Hinton, G., Vinyals, O., \& Dean, J. (2015)\cite{hinton2015distilling}, disappear from the full formula suggested by same authors. Then, we tested the same $Res16Net34C_{Half}$ model with $\alpha = 0.5$, to understand the effect of the same classic Knowledge Distillation method on the 3D Semantic Segmentation task.

We purpose a new loss function, which is mostly compatible with relatively smaller networks. To be more detailed,  we suggest using Upsampling convolution over the Feature Map of student network layers, which is transformed into the same dimension as the teacher model. Figure \ref{fig1}, it is shown 2 blocks, which transform the student Encoder and Decoder last layer feature maps, to the same dimensionality as the teacher. Spatio Temporal Convolution layer is used during this step with kernel size 1, which plays an Upsampling role for a number of neurons on the feature map layer. For the $Res16Net34C_{Half}$ model, we are using 128 -> 256 for the Encoder feature map, and 48->96 for the Decoder feature map layers, as Convolution input -> output.
This method, is similar to the Bottle Neck effect, like in the paper by Tishby, N., \& Zaslavsky, N. (2015, April).\cite{tishby2015deep}, that forces the student network feature map last layer to learn the same information as teacher respective one. For example for the $Res16Net34C_{Half}$ model, we make encoder 128 Sparse Convolutional block to represent the same information as the respective  $Res16Net34C$ 256 block. We have done the same technique for Encoder's last Feature map layer, by Upsampling 16->32(See \ref{fig1}). This method is similar to the Dimensionality reduction method(see examples by Van Der Maaten, L., Postma, E., \& Van den Herik, J. (2009) \cite{van2009dimensionality}). A similar idea is also used in AutoEncoders and there is also a paper by Wang, Y., Yao, H., Zhao, S., \& Zheng, Y. (2015)\cite{wang2015dimensionality}, which proves, that in autoencoder networks, dimensionality reduction effect is more significant. 

Another assumption of the new layer applying for the loss function is making the training process faster. During the backpropagation process, when the model output is taken not only from the last layer but also from the intermediate layers, we are making the gradients relatively small. This is because, the gradient of encoder feature map loss, is only included before the encoder network, while the gradient of the last layer includes the encoder + decoder network while passing through the neural network and backpropagating via chain rule. This problem is described in numerous papers, such as the paper by Hochreiter, S. (1998) \cite{hochreiter1998vanishing}

\section{Experiments and Results}
\label{exp}
In this section, we present our experiments and results of different loss functions for Knowledge Distillation on 3D Semantic Segmentation. We evaluate our results on the Scannet V2 dataset\cite{dai2017scannet}. We have used both standard loss functions purposed by  Hinton, G., Vinyals, O., \& Dean, J. (2015)\cite{hinton2015distilling}, and also our suggested loss functions(explained on \ref{loss}. Our results show that by using Knowledge Distillation methods, we can get similar results as state-of-the-art models, by significantly decreasing model parameters and making inference faster as well. We also show, that by using feature map loss functions, we increase the performance and make training more stable and faster. For background, we suggest looking at \ref{sec2}. 
\subsection{Model selection}
For better knowledge distillation tasks, a really good teacher model should be selected. To do so, we have selected $Res16Net34C$ architecture suggested by MinkowskiNets\cite{choy20194d} as the teacher model and trained it with an augmentation technique improved by Mix3D\cite{nekrasov2021mix3d}. For student network, we selected $Res16Net34C_{Half}$ and $Res16Net34C_{Quarter}$ architectures(explained on  \ref{model}). During this step, we decreased our model sizes with the following table shown below \ref{modelsize}.
\begin{table}[h]
\centering
\caption[width=4cm]{\\The validation results are done on Scannet V2 dataset.}

\begin{tabular}{|c|c|c|c|}
\hline
\textbf{Model} & \textbf{Number of Parameters(millions)} & \textbf{Val mIoU}\\ \hline
$Res16Net34C$ & 39.7m & 69\%\\ \hline
$Res16Net34C_{Half}$ & 9.6m & 66.4\%\\ \hline
$Res16Net34C_{Quarter}$ & 2.3m & 60.6\%\\ \hline
\end{tabular}

\label{modelsize}
\end{table}

As you can see, the model parameters decreased significantly (around 4 times each step). We show around a 2.6\% difference of mean Intersection over Union between models trained with $Res16Net34C_{Half}$ and $Res16Net34C$ while making a 4 times smaller network with the same architecture. Our model parameter reductions are done, by symmetrical reducing the number of kernels on each block of Sparse Convolutions, by 2 and 4 times respectively for $Res16Net34C_{Half}$ and $Res16Net34C_{Quarter}$ models. This means, a reduction of model parameters with $N^2$, which is 4 and 16 times smaller model. To evaluate this approach, we have used different loss functions over the same network architectures, to analyze the performance of models, on train and validation Datasets. Model evaluations are done on the Scannet V2 dataset, by using training and validation split.
For more details(per class) analysis, see \ref{super}.

\subsection{Dataset}
During the analysis, the Scannet V2 dataset, which contains rich and diverse 3D semantic segmentation information, was employed as a benchmark to systematically evaluate and compare the performance of various Knowledge Distillation methods. This approach ensured a comprehensive understanding of the effectiveness and potential of each method in the context of 3D deep learning models.
For more details, the Scannet V2 dataset is a large-scale dataset for 3D scene understanding and reconstruction. It contains a collection of RGB-D images, 3D point clouds, and semantic labels for indoor scenes captured using commodity depth sensors. The dataset was introduced by Dai, A., Chang, A. X., Savva, M., Halber, M., Funkhouser, T., \& Nießner, M. (2017)and has since been widely used in research for tasks such as semantic segmentation, instance segmentation, and 3D reconstruction.

The dataset consists of 1513 scans, each containing a set of RGB-D images and corresponding 3D point clouds, captured by an RGB-D sensor such as the Microsoft Kinect. The scans cover a variety of indoor scenes such as apartments, offices, and public spaces, with different furniture, decorations, and layouts. The dataset also provides semantic classes for each point in the point cloud(with a total count of 20), including labels such as wall, floor, bath, chair, table, and so on. The annotations are provided at three levels of detail: coarse, fine, and instance-level.

\subsection{Stable training}

We suggest using additional loss functions, by using Upsampling layer over Encoder/Decoder feature maps. This way, we have more control over network training and what the network learns. In analysis, we found out, that as long as you increase control over the network, the training process becomes much more stable. For example, in the first iterations of student networks, with the architecture of  $Res16Net34C_{Half}$, we show that the training of networks is much more stable when we include Feature Map loss(explained on \ref{loss}. It is due to the fact, that we force the model to mimic the feature map space, by making student layers Bottle Neck. Hence, the model tries to minimize loss, by moving directly to the feature map. See the example in the table below.

\begin{figure}[hbt!]
    \centering
    \includegraphics[height=40mm]{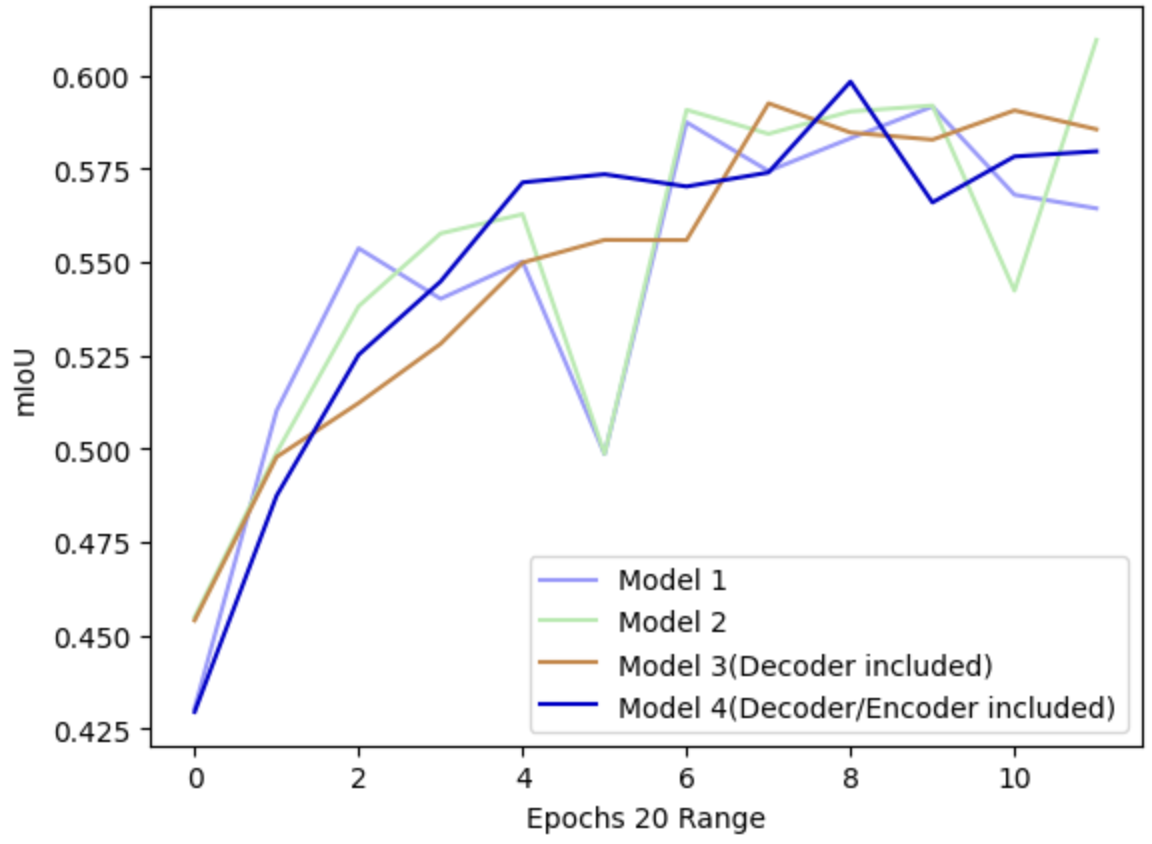}
    \captionsetup{width=.6\linewidth}
    \caption[width=4cm]{First 200 Epochs that run on our tested 4 models (that are from Res16UNet34C\_Half architecture). It shows that by adding additional losses to keep in control the distillation process, we make the training process more stable at the beginning of the process.}
    \label{figRun}
\end{figure}
The idea of making training more stable, allows us to make a much higher learning rate, which itself allows faster training with a higher learning rate. A similar idea has been used by Gao, M., Shen, Y., Li, Q., Yan, J., Wan, L., Lin, D., ... \& Tang, X. (2018).\cite{gao2018embarrassingly}, which makes the training stable, by giving weights to the classes of the network.

\subsection{Supplementary Materials}
During our analysis, we included both original and Replicated results of MinkowksiNet\cite{choy20194d} with data augmentation of Mix3D\cite{nekrasov2021mix3d} method. Our analysis here is done on Scannet V2 dataset\cite{dai2017scannet}. We separated Supplementary Materials sections for our research, that is describing details about the training process, and also details of our experiments.

\subsection{Training (Supplementary Materials)}



To apply different losses suggested by us (see section \ref{loss}), we implemented the hydra config file, which contains all loss functions we applied. Those combinations are explained in the \href{https://github.com/madanela/smaller3d/tree/main/config_script}{ConfigCombinations} section. We also included detailed training information in Table \ref{bigtable}, which provides details about hyperparameters $\alpha$ and T.

Our experiments were conducted using a single 1080 8GB GPU, providing sufficient computational power for our training needs. Each training session took around four days to complete. We utilized the CUDA 10.1 library and the PyTorch framework for our experiments. To be more specific, we used MinkowksiNet\cite{choy20194d}, which is an engine designed specifically for 4D-SpatioTemporal ConvNets and is a key component of MinkowksiNet\cite{choy20194d} and Mix3D\cite{nekrasov2021mix3d}.

In addition to the mentioned points, we used the StepLR learning rate scheduling technique, which adjusts the learning rate at specified intervals, leading to more efficient convergence. Following the standard evaluation metrics from previous works, we used mean Intersection over Union (mIoU) and mean Accuracy (mAcc) for evaluation.
\subsection{Additional Information (Supplementary Materials)}
\label{super}

The experiments are done with a fixed T value (suggested by \cite{hinton2015distilling}). Also, we have changed the $\alpha$ parameter on our first experiment, by first picking it 1, which means we use only standard cross entropy loss over label and student network predictions. Then, in our further experiments, we used $\alpha = 0.5$, which proved to be better, as suggested by \cite{hinton2015distilling}. The results of this, are included on \ref{bigtable} first and second student network results. 

This section also includes per-class analysis. As demonstrated, our last model, based on our proposed loss with Encoder and Decoder feature map (explained in section \ref{loss}) and model (explained in section \ref{model}), performs better in some cases (for example, picture or bed) than the four times larger teacher model Res16Net34C [2]. However, in some cases, the student model fails to replicate similar results to the teacher in representing 7 specific classes, and this is done by a relatively large margin (cases like sink (by 8\%), and shower\_curtain (by 6\%)).

\label{bigtable}
\begin{table}[hbt!]
  \centering
  \caption{Measurements of models by validation and different class mIoU}
    \small
    \footnotesize\setlength{\tabcolsep}{1pt}
    \begin{tabular}{l@{\hspace{1pt}} *{8}{|c}}
    \hline
    \multirow{1}{*}{Models} & \multicolumn{2}{|c|}{Teacher Networks} & \multicolumn{5}{|c|}{Student Networks \ All are based on Res16UNet34C}\\
    \cline{2-8}
    & Mix3D & Replicated & Half & Half & Half & Half & Quarter& \\
    \hline 

    voxel(5cm) & \checkmark & \checkmark  & \checkmark & \checkmark & \checkmark & \checkmark & \checkmark     \\
    \hline
    $\alpha$ & \xmark & \xmark  & 1 & 0.5 & 0.5 & 0.5 & 1 &   \\
    \hline 
    $T$ & \xmark & \xmark  & 1 & 1 & 1 & 1 & 1 &  \\
    \hline 
    $Decoder Loss$ & \xmark & \xmark & \xmark  & \xmark & \checkmark & \checkmark & \xmark   &  \\
    \hline 
    $Encoder Loss$ & \xmark & \xmark & \xmark  & \xmark & \xmark & \checkmark & \xmark   &  \\
    \hline 
    mIoU & \textbf{69.1\%}& 69\% & 65.9\%  & 66.3\% & 66\% &66.4\% & 60.6\% & \\ 
    \hline
    By Class &  & &  &  &  &  &  &\\
    \midrule
    bathtub & \textbf{81.1\%} &80\% & 79.5\%  & 79.3\% & 77.2\%&79.7\%& 74.1\% & \\
    bed & 79.2\% & \textbf{80.2\%}& 78.8\%  & 78.9\% & 78.2\% &79.6\%& 76\% & \\
    bookshelf & 78.1\% &\textbf{78.4\%}& 72.9\%  & 77.3\% & 75.8\% &75.8\%& 71.6\% & \\
    cabinet & 63.9\% & \textbf{64\%} & 61.0\%  & 63.7\% & 61.4\%&63.3\%& 56.4\% & \\
    chair & 89.2\% & \textbf{90\%} & 87.1\%  & 86.9\% & 87.3\%&88\%& 83.4\% & \\
    counter & \textbf{62.2\%} &58.1\%& 59.3\%  & 59.5\% & 55.5\%&61.3\%& 54.8\% & \\
    curtain & \textbf{67.0\%} &64\%& 60.2\%  & 61.2\% & 58.7\%&60.0\%& 54.2\% & \\
    desk & 62.1\% & \textbf{63.8\%} &60.1\%  & 62.7\% & 59.3\%&62.5\%& 56.7\% & \\
    door & 59.3\% & \textbf{60\%} & 55.4\%  & 54\% & 55.4\%&55.5\%& 49.8\% & \\
    floor & \textbf{93.9\%} & \textbf{93.9\%} & 93.5\%  & 93.6\% & 93.5\%&93.7\%& 92.7\% & \\
    otherfurniture & 54.3\% & \textbf{56.8\%} & 56.3\%  & 55.4\% & 55.5\%&55.8\%& 45.7\% & \\
    picture & 25.6\% & 25\% & 23.9\%  & 24.3\% & 24.3\%&\textbf{26.1\%}& 16.1\% & \\
    refridgerator & 54.1\% & \textbf{54.7\%} & 43.2\%  & 44.8\% & 48.2\%&51.4\%& 40.7\% & \\
    shower\_curtain & \textbf{64.0\%} & 63.8\% & 62.4\%  & 61.9\% & 61.5\%&58\%& 47.7\% & \\
    sink & 64.1\% &\textbf{64.3\%} & 57.1\%  &54.3\% & 56\%&56\%& 46.4\% & \\
    sofa & \textbf{81.1\%} & 80\% & 77.5\%  & 77.2\% & 78.1\%&77.8\%& 74.1\% & \\
    table & 70.8\% & \textbf{71.8\%} & 71.1\%  & 70.9\% & 70.4\%&71.5\%& 66.1\% & \\
    toilet & 89.8\% & \textbf{90.4\%} & 83.6\%  & 85.3\% & 86.9\%&85\%& 79\% & \\
    wall & 82.1\% & \textbf{82.7\%} & 80.6\%  & 80.5\% & 80.5\%&81\%& 77.4\% & \\
    window & \textbf{59.8\%} & 58.5\%  & 55.6\%  & 54.4\% & 54.6\%&54.7\%& 50.5\% & \\
    \midrule

    \hline
  \end{tabular}
  \label{tab:table} 
\end{table}

This analysis is done using a 5cm voxel size. We show, that in some cases(for example picture, or bed) our suggested $Res16Net34C_{Half}$ model performs better than the 4 times bigger teacher model $Res16Net34C$\cite{choy20194d}. However, in some cases, the student model fails to replicate similar results to a teacher, to present specific classes, and this is done by a relatively large margin(cases like a sink(by 8\%), and shower\_curtain(by 6\%).

\section{Conclusion}

In this paper, we have presented the Knowledge Distillation technique for 3D deep learning, specifically for the 3D Semantic Segmentation task. By employing this technique, we aimed to reduce the size of state-of-the-art models while maintaining their performance, resulting in computationally less expensive and faster inference.

Our experiments with various model architectures, such as Minkowski Engine\cite{choy20194d} and Mix3D\cite{nekrasov2021mix3d}, demonstrated the effectiveness of our approach. We successfully reduced the number of neurons in each layer by a factor of 4 and 16, while still approximating the performance of the original models. We also proposed new loss functions, which provided additional control over the distillation process and led to more stable and faster training.

Our results show that Knowledge Distillation techniques can be effectively applied to 3D Semantic Segmentation tasks, providing a viable solution for reducing the model size and computational requirements without sacrificing performance. Furthermore, our proposed loss functions, which utilize Upsampling over encoder/decoder feature maps, offer an additional level of control that contributes to increased training stability and faster convergence.

In conclusion, our work demonstrates the potential of Knowledge Distillation for 3D deep learning tasks and provides a foundation for future research in this area. By making current models smaller and more efficient, we hope to enable the deployment of advanced 3D deep learning techniques in real-world applications with limited computational resources.

\section*{Acknowledgments}
AUA (American University of Armenia) provided hardware resources for the project. I would like to acknowledge the support and assistance in particular David Davidian. The article solely reflects the authors only (not any organization).

\bibliographystyle{unsrt}  
\bibliography{references}  

\newpage

\end{document}